# Designing Artificial Cognitive Architectures: Brain Inspired or Biologically Inspired?

Emanuel Diamant
VIDIA-mant, Kiriat Ono, Israel
emanl.245@gmail.com

**Abstract:** Artificial Neural Networks (ANNs) were devised as a tool for Artificial Intelligence design implementations. However, it was soon became obvious that they are unable to fulfill their duties. The fully autonomous way of ANNs working, precluded from any human intervention or supervision, deprived of any theoretical underpinning, leads to a strange state of affairs, when ANN designers cannot explain why and how they achieve their amazing and remarkable results. Therefore, contemporary Artificial Intelligence R&D looks more like a Modern Alchemy enterprise rather than a respected scientific or technological undertaking. On the other hand, modern biological science posits that intelligence can be distinguished not only in human brains. Intelligence today is considered as a fundamental property of each and every living being. Therefore, lower simplified forms of natural intelligence are more suitable for investigation and further replication in artificial cognitive architectures.

## 1. Introduction

Pursuing the task of intelligent machines design, thinking about their possible architectures, we always look first at the prototypes that the nature has prepared for us so plentiful and generously. Most commonly, the human brain is chosen to be the source of our inspiration.

This way of thinking – considering the human brain as the source of our inspiration - was established more than 50 years ago at the Dartmouth college meeting (in summer of 1956), where the concept of Artificial Intelligence has been proclaimed and inaugurated [1].

What is "intelligence" was not defined at the meeting, but it was self understood that human intelligence is what's being meant and the human brain is supposed to be its most probable location. Soon after that, the Artificial Neural Network (ANN) toolkit was devised and put in use for AI studies and investigations. The ANN was contrived as a collection of small interconnected computational units (called artificial neurons), which are supposed to imitate the biological neurons of the human brain, and in a greatly simplified form simulate the way in which the brain supposedly performs its duties.

## 2. The Raise and Fall of AI

During the following 50 years, the ANNs have seen several ups and downs. Nowadays, we witness an extreme rise in Deep Learning Neural Nets (DLNNs) proliferation accompanied with an unprecedented wave of promotion and advertising. Actually, the excitement around DLNNs is not entirely unjustified: In recent years, it has turned out that they have emerged as a powerful tool for a variety of application in a range of domains, such as language and speech modeling, image and video analysis, protein structure prediction, and so on [2].

However, despite of the hype, the impact of the ANN (AI) techniques on human brain research and brain functionality modeling remains restricted and dim. There is a known and a widely recognized latent deficiency in the ANNs applications: the ANN-DLNN designers have a critical lack of understanding of how their mechanisms work. Neural networks are considered black boxes because it is impossible to grasp how they make their work done [3]. Their internal operational principles remain unclear and ambiguous. And there is a persistent lack of knowledge on why and how they achieve their impressive performances. From a scientific point of view, that is absolutely unsatisfactory. Without a clear understanding of how and why the DLNNs work, the development of better models inevitably reduces to trial-and-error experimentation. For that reason, contemporary Artificial Intelligence R&D looks more like a Modern Alchemy enterprise rather than a respected scientific or technical undertaking.

## 3. The Translational Neurobiology advent

Despite the dominance of the AI doctrine, which presumes that human beings are the only possible carriers of natural intelligence, the new wave of biological scientists has adopted a different, separate view on the subject. Dabbed as The Translational Neurobiology (TN) approach it posits that the brain's functional structure can be dissected into more basic constituting components. And after a detailed analysis of segregated fragments some of them could be retranslated back into the human brain and to be reintegrated therein for further brain impairments care and treatment [4]. The reverse information flow (from the clinical system to the research installation) is also a very important issue in the TN context. It allows to grasp insights about the cognitive mechanisms that support and enable behavior of the investigated living creatures. Thus, fundamental questions about brain machinery and working can be explored and clarified.

What is common to the AI and the TN paradigms is that the term "intelligence" is used without any attempt to define or elucidate it. That does not embarrass in any way the TN designers: they use a very close substitute for "intelligence" – a not less ambiguous and uncertain term "cognition" (cognitive). That is a widespread and common practice (today), so we will also keep for us the right to use them (the terms "intelligence" and "cognition") interchangeably.

What AI and TN approaches do not share benevolently is the idea that intelligence (cognition) is an exceptionally human trait. Indeed, today, a growing part of the research community is ready to speak about: 'eukaryotic intelligence', 'plant intelligence', 'fish intelligence', 'animal intelligence', and so on [5], [6]. A similar list of examples can be provided for the "cognitive" swap for "intelligence": 'cognitive bacteria', 'zebrafish cognition', 'cognitive abilities of birds', 'animal cognition' [7], [8]. To sum up, today consciousness, cognition and intelligence (that are frequently used interchangeably) are firmly grounded fundamental concepts associated with life and living beings [9].

Investigation into primitive cognitive features revealed in low-level biological creatures led to an impressive advancement in low-level cognitive-intelligent-conscious machinery studies. Projecting these discoveries to human cognitive abilities examination has immediately put forward the whole brain research programme. TN technique is acknowledged as a crucial factor of this achievement. But, AI research can also benefit from these innovations – it can now leave the task of a whole brain investigation in favor of a more traceable composite low-level cognitive feature exploration and delineation.

## 4. Why the hell you are still unhappy

Among the general fun, one perplexing question remains unsolved: the existing diversity and multiplicity of cognition-intelligence forms (slime, fish, plant, animal, human, and so like "cognitions") request a reasonable explanation what makes them so different and yet so alike in many ways. A possible explication can be as such: cognition (intelligence) is not a state of mind – it is a process. Neuron structures and nervous systems characteristic for higher organisms are absent in low-level living creatures. However, cognitive functions are conserved across the entire spectrum of living species. In such a case, the question is inevitable: what the hell they all are processing? And the answer is also pretty well known: The human brain is processing information! That is a very common and a widely accepted dictum. Therefore, we can extend it to the entire cognition-intelligence domain: **Cognition (intelligence) is the ability to process information** [10].

That leads immediately to the next bewildering question: What is information?

There is a widespread conviction that a consensus definition of Information does not exist, and considering the multitude of its applications, it may be not even supposed to exist. I do not agree with this convention. On several occasions, I have already published my opinion on that subject [11], [12], [13]. This time, with all fitting excuses, I would like to repeat some fractions of these previous publications (in order to preserve the consistency of our discussion).

Relying on Kolmogorov's 1965 year paper [14], I have developed my own definition of information that can be articulated in the following way: "**Information is a linguistic description of structures observable in a given data set**".

To make the scrutiny into this definition more palpable I propose to consider a digital image as a given data set. A digital image is a two-dimensional set of data elements called picture elements or pixels. In an image, pixels are distributed not randomly, but, due to the similarity in their physical properties, they are naturally grouped into some clusters or clumps. I propose to call these clusters **primary or physical data structures**.

In the eyes of an external observer, the primary data structures are further arranged into more larger and complex agglomerations, which I propose to call **secondary data structures**. These secondary structures reflect human observer's view on the grouping of primary data structures, and therefore they could be called **meaningful or semantic data structures**. While formation of primary (physical) data structures is guided by objective (natural, physical) properties of the data, the subsequent formation of secondary (semantic) data structures is a subjective process guided by human conventions and habits.

As it was said, **Description of structures observable in a data set should be called "Information".** In this regard, two types of information must be distinguished – **Physical Information and Semantic Information**. They are both language-based descriptions; however, physical information can be described with a variety of languages (recall that mathematics is also a language), while semantic information can be described only by means of a natural (human) language. (More details on the subject could be found in [15]).

Those, who will go and look in [15], would discover that every information description is a top-down-evolving coarse-to-fine hierarchy of descriptions representing various levels of description complexity (various levels of description details). Physical information hierarchy is located at the lowest level of the semantic hierarchy. The process of sensor data interpretation is reified as a process of physical information extraction from the input data, followed by an attempt to associate this physical information (about the input data) with physical information already retained at the lowest level of the semantic hierarchy.

If such an association is attained, the input physical information becomes related (via the physical information retained in the system) with a relevant linguistic term (see also the block diagram in [11]), with a word that places the physical information in the context of a phrase, which provides the semantic interpretation of it. In such a way, the input physical data object becomes named with an appropriate linguistic label and framed into a suitable linguistic phrase (and further – in a story, a tale, a narrative), which provides the desired meaning for the input physical information. (Again, more details can be found on the website [15]).

## 5. What follows from this information

From this definition of information follows, first, that Intelligence (Cognition, Consciousness), which are the product (a result) of information processing, can be of various and different levels of complication that depend only on the complexity and the volume of information that is supposed to be processed. Second, the means required for processing different levels of information have to be adequate to the information size and grade.

It must be also taken into account that Intelligence (Cognition, Consciousness) are tightly related with semantic information processing, often called "cognitive information" processing. As we know now, semantic information is carried out as pieces of text, storytelling fragments, and narrative structures. How semantic information processing is realized in different living beings is not known to us yet. But what we do know certainly is that this is a linguistic texts processing procedure, which contemporary data processing computers don't supposed to handle yet.

A very important feature is information duality, which ensues from the above given information definition. What follows from it is that semantic information cannot be derived from the data-driven physical information. Data and information don't have casual relations, they are complementary, not causal. Contemporary information processing machines are repeatedly stumble upon this hurdle. That is the most prevalent misconception governing the today's artificial Cognitive Architecture (CA) design mainstream.

## 6. Some concluding remarks

So, the question in the title "Brain inspired? Or biologically inspired?" is not so stupide as it may look at the first glance. Brain inspired approach, usually implemented as ANN (DLNN) arrangement, turns out as a misleading dead-ended solution. And the only way to push further our CA designs is to deepen our understanding about the nature of information and the mysteries of natural information processing paradigms.

I understand that proposed here Information processing paradigm is hard to swallow, and I imagine that you will not hurry to apply it in your future CA designs. Therefore, before I give up and finish my contribution, please allow me to reveal to you some of the paradigm's hidden charms, which become clear and evident just recently.

The productive power of the proposed approach is illustrated here by the new findings revealed in the latest biologically inspired investigations. In the ENEURO journal, published on May 14, 2018, an UCLA research group reported a successful memory transfer from one marine snail to another [22].

The experiment reliably confirms my assumption that Information (and all its derivatives – thoughts, feelings, memories) are material entities, that is, are strings of nucleotides comprising the text of an information message. As such, they can be processed, manipulated, and even relocated.

Memories are not more arrangements of adjustable synapses, as the CA designers traditionally view and exploit them. Memories are real linguistic descriptions of observed structures, that we retain and recycle in our brains and our CA designs.

Therefore, the time is ripe to move from Brain Inspired Cognitive Architectures to Biologically Inspired Cognitive Architectures.